\documentclass[11pt,a4paper]{article}
\usepackage[hyperref]{acl2021}
\usepackage{times}
\usepackage{float}
\usepackage{latexsym}
\usepackage{spverbatim}
\usepackage{fancyvrb}
\usepackage{listings}

\usepackage{microtype}
\usepackage{multirow}
\usepackage{graphicx}
\usepackage{array}
\usepackage{microtype}
\usepackage{multirow}
\usepackage{graphicx}

\aclfinalcopy 

\definecolor{deepblue}{rgb}{0.16, 0.32, 0.75}
\definecolor{indiagreen}{rgb}{0.00, 0.44, 0.00}

\title{Summary Grounded Conversation Generation}
\author{Chulaka Gunasekara, Guy Feigenblat, Benjamin Sznajder, Sachindra Joshi, \\ {\bf David Konopnicki}
 \\
        IBM Research AI\\
\texttt{chulaka.gunasekara@ibm.com, \{guyf, benjams\}@il.ibm.com}\\ \texttt{\{jsachind@in, davidko@il\thanks{~~Current address:   david.konopnicki@booking.com}~\}.ibm.com}}

\date{}

\begin{document}
\maketitle
\begin{abstract}

Many conversation datasets have been constructed in the recent years using crowd-sourcing. However, the data collection process can be time consuming and presents many challenges to ensure data quality. Since language generation has improved immensely in recent years with the advancement of pre-trained language models, we investigate how such models can be utilized to generate entire conversations, given only a summary of a conversation as the input. We explore three approaches to generate summary grounded conversations, and evaluate the generated conversations using automatic measures and human judgements. We also show that the accuracy of conversation summarization can be improved by augmenting a conversation summarization dataset with generated conversations.


\end{abstract}

\section{Introduction}
\begin{figure*}[t]
    \centering
    \includegraphics[width=0.78\textwidth]{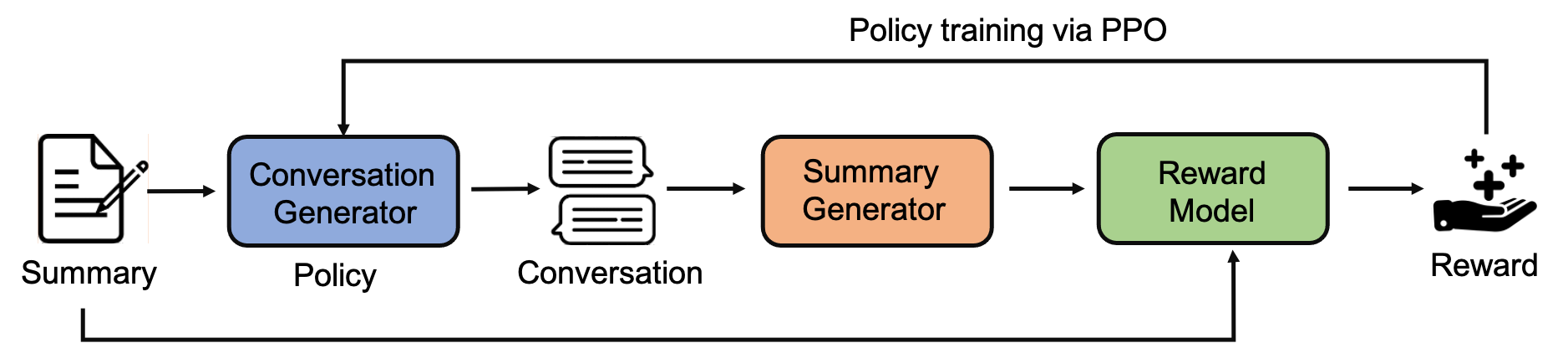}
    \caption{The RL based conversation generation framework}
    \label{fig:RL_train}
\end{figure*}
Automatic conversation systems require large quantities of data to learn task specific language patterns and underlying conversation policies. Such data either come from human-to-human conversation logs \cite{lowe2015ubuntu, hardalov2018towards} or is collected in crowd-sourced environments, where two or more crowd-workers play specific roles under some guidelines 
\cite{zhang2018personalizing, budzianowski2018multiwoz}. Since real human-to-human conversation logs are scarce, many datasets have been created using the latter approach. However, crowd-sourced conversation data collection is time consuming, costly and presents multiple challenges to ensure data quality \cite{kang2018data}.

Conversation summarization is an emerging research area that has been ill-studied due to the lack of large-scale datasets. Most existing public datasets in this domain are small, for example, AMI meeting corpus \cite{mccowan2005ami} contains $137$ summary transcripts. CRD3 \cite{rameshkumar2020storytelling} is a spoken conversation dataset that consists of $159$ conversations and summaries. Samsum \cite{gliwa2019samsum}, the only large scale dataset for conversation summarization, contains over $16,000$ open-domain conversations and summaries created artificially by humans. 

Large scale pre-trained language models (PLMs) \cite{lewis2020bart, brown2020language, raffel2020exploring} have been used in various text generation tasks \cite{budzianowski2019hello, min2020ambigqa, cachola2020tldr}. In recent studies, PLMs are used to generate training data for natural language processing (NLP) applications. For example,~\citet{anaby2020not, yang2020g} use PLMs to create paraphrases for intent classifiers in conversation systems, and show that, when the original datasets are augmented with the generated data, performance improves. More recently~\citet{mohapatra2020simulated} generated entire conversations grounded on instructions that are provided to crowd-workers using a modular approach, where different PLMs are trained for different roles.


\textbf{Our Contributions: }
We investigate how PLMs can be utilized to generate  entire conversations that are grounded on a given summary. We explore three approaches:
(1) Supervised Learning (SL) based conversation generation \textit{(SL-Gen)}: where, a PLM is trained to generate an entire conversation, taking the summary of a conversation as input, (2) Reinforced Learning (RL) based conversation generation \textit{(RL-Gen)}: where, we further improve the \textit{SL-Gen} method using the quality of the generated conversations as a reward, and (3) Controlled turn-by-turn conversation generation \textit{(CN-Gen)}: which allows us to generate conversations turn-by-turn, constrained on the summary and a set of pre-defined control parameters. 
We evaluate the quality of the generated conversations by conducting  automatic and human evaluation. We also show that once a conversation summarization dataset is augmented with the generated conversations, the performance of the downstream summarization task is improved.

\section{Summary grounded conversation generation} \label{sec:methods}

In the conversation summarization task, a model takes a conversation as  input, and learns to generate a summary. We study the inverse of that problem, where the input to our model is a summary, and the model generates a conversation. In this section, we propose three models for this task and the hyper-parameters used in training the models are available in Section A of the appendix. 
%




\subsection{SL based generation (SL-Gen)}
A seq2seq model can be trained for this task by providing a summary as the input and generating a conversation token-by-token. As PLMs have shown significant improvement over the traditional seq2seq architecture for text generation, we use a GPT-2 model and fine-tune it to generate a conversation given a summary as the input. Our input to the model follows the following format: {\small \textsl{$<$bos$>$summary text $<$dialog$>$conversation text$<$eos$>$}}. We also use different token-type-ids to indicate the summary and the conversation text. The model is trained to optimize Cross Entropy loss. 

\subsection{RL based generation (RL-Gen)}
Many studies train text generation models with RL \cite{paulus2018deep, li2016deep}, where the generator network is optimized with a task specific reward. We investigate how the quality of the generated conversation can be used as a reward to improve the generation network. To this end, we train a summary generator network, which generates a summary, given a conversation. We measure the quality of the generated conversation by identifying the similarity between the summary of the generated conversation (generated, in turn, by the summary generator network) and the ground truth summary. The similarity score is used as a reward to train the conversation generation model. Our RL based generation framework is shown in Figure \ref{fig:RL_train}, and the critical components are described below. \\
\textbf{Conversation Generator:}  A trained SL-Gen model is used as the conversation generator, which, given an summary can generate a conversation.\\
\textbf{Summary Generator:} We use a lightweight variant of BART \cite{lewis2019bart}, named \textit{DistilBART}, which is fine-tuned on the 
Extreme summarization task \cite{narayan2018don}.
We further fine-tune this instance on the conversation summarization data by providing the conversations as the input and training the model to output summaries.\\
\textbf{Reward Model:} Once the Summary Generator generates an output summary for the generated conversation, the reward model compares it with the ground truth summary, which was used to ground the conversation generation. As \citet{paulus2018deep} we use ROUGE-2 F1-score
as the reward.\\ 
\textbf{Policy training:} We use proximal policy optimization 
\cite{schulman2017proximal} as the optimizer for the policy training as it prevents the generator from deviating far away from the pretrained LM \cite{wu2020textgail}. 

\subsection{Controlled conversation generation}
We propose another approach, \textit{(CN-Gen)}, for conversation generation, which grants more control over the properties of the generated conversations. Here, we generate one utterance of the conversation at a time, as opposed to the \textit{RL-Gen}, where we generate the whole conversation at once. The properties of the generated conversations is controlled by adding several components to the input sequence to the model. The following three variables were used as the control parameters, (1) Number of remaining turns to generate in the conversation (Num turns): During the generation of a turn, we indicate the remaining number of turns in the conversation. In generating a $n$ turn conversation, this starts with $n$ for the first turn and reduces by $1$ after the generation of each turn, (2) The speaker of the next turn (Speaker): This indicates to the model the speaker of the next turn, and (3) The length of the next turn (Turn length): We define, $3$ categories of lengths: Short ($\leq$ 3 tokens), Long ($>$ 10 tokens) and Medium (otherwise). 


\begin{figure*}[ht]
    \centering
    \includegraphics[width=0.9\textwidth]{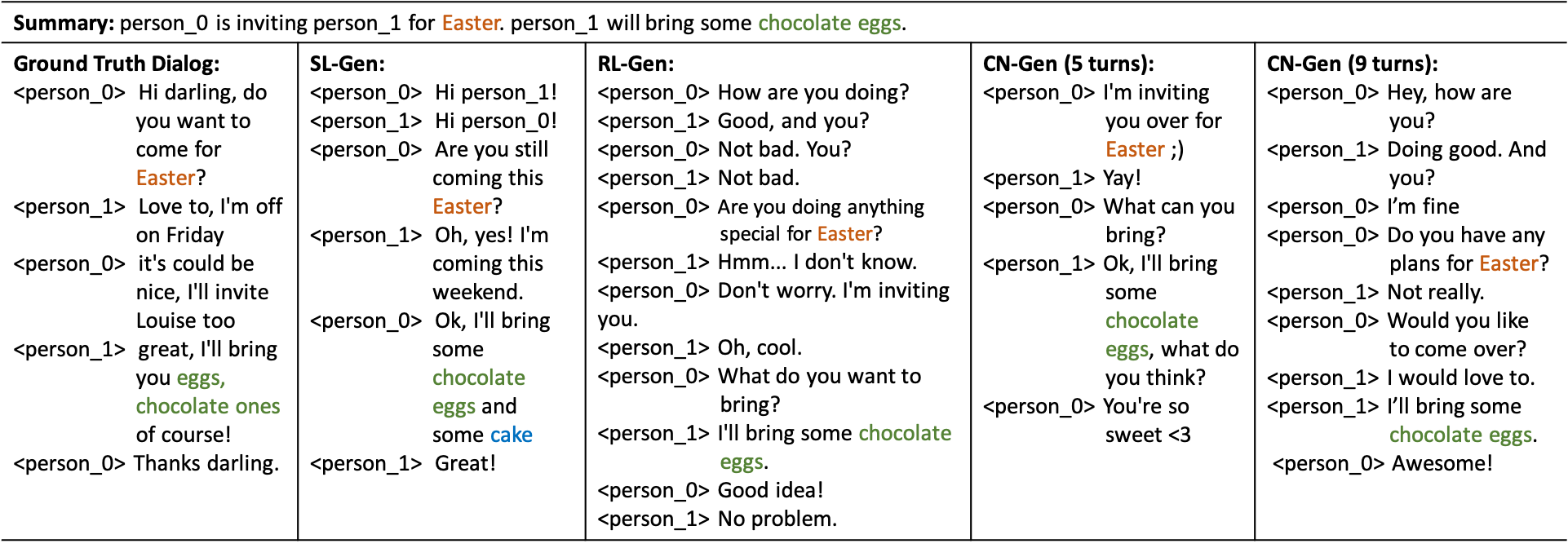}
    \caption{Examples of a conversations grounded on the same summary. The key terms are highlighted in colors.}
    \label{fig:example_convs}
\end{figure*}

We use the following input representation to fine-tune a GPT-2 model: {\small \textsl{$<$bos$>$ summary text $<$context$>$ dialog context  $<$turns\_to\_go$>$ Num turns $<$speaker$>$ speaker $<$turn\_length$>$ turn length $<$turn$>$ utterance $<$eos$>$}}. Changing these parameters allows us to generate different variants of conversations which are grounded on the same summary. During training, we obtain the values for the control parameters from the ground truth conversations, and at inference we randomly select the next speaker, number of turns of the conversation to be generated (in a range of 4-15 turns), and the next turn length. In Table \ref{table:cn_gen_samples} we show conversations of different lengths that were generated by the CN-Gen approach grounded on the same summary by changing the control parameters.

\begin{table}[t]
\centering
\begin{tabular}{p{100pt}|p{100pt}p{100pt}p{100pt}}
   \hline
    \multicolumn{4}{l}{\tiny{\textbf{Summary:} person0 will be late. person1 will order pasta with salmon and basil for her.}}\\
   \hline
   
   \begin{tabular}[t]{p{.20\linewidth}p{.60\linewidth}}
   \multicolumn{2}{l}{\tiny{\textbf{2 turn conversation:}}}\\[-4 pt]
        \tiny{$<$Person0$>$} & \tiny{I'll be late}\\[-7 pt]
        \tiny{$<$Person1$>$} & \tiny{I'll order some pasta with salmon and basil for you.}\\[-7 pt]
   \end{tabular}

   & 
   \begin{tabular}[t]{p{.20\linewidth}p{.60\linewidth}}
   \multicolumn{2}{l}{\tiny{\textbf{3 turn conversation}}}\\[-4 pt]
        \tiny{$<$Person0$>$} & \tiny{I'll be late.}\\[-7 pt]
        \tiny{$<$Person1$>$} & \tiny{I'll order some pasta with salmon and basil for you.}\\[-7 pt]
        \tiny{$<$Person0$>$} & \tiny{Thanks a lot!}\\[-7 pt]
   \end{tabular}

    & 
\\\hline

   \begin{tabular}[t]{p{.20\linewidth}p{.60\linewidth}}
   \multicolumn{2}{l}{\tiny{\textbf{6 turn conversation}}}\\[-4 pt]
        \tiny{$<$Person0$>$} & \tiny{Hello, I am going to be late.}\\[-7 pt]
        \tiny{$<$Person1$>$} & \tiny{Ok}\\[-7 pt]
        \tiny{$<$Person1$>$} & \tiny{I'll order some pasta with salmon and basil}\\[-7 pt]
        \tiny{$<$Person0$>$} & \tiny{Ok, sounds good!}\\[-7 pt]
        \tiny{$<$Person0$>$} & \tiny{Thank you!}\\[-7 pt]
        \tiny{$<$Person1$>$} & \tiny{No problem}\\[-7 pt]
   \end{tabular}

   &
   
   \begin{tabular}[t]{p{.20\linewidth}p{.60\linewidth}}
   \multicolumn{2}{l}{\tiny{\textbf{10 turn conversation}}}\\[-4 pt]
        \tiny{$<$Person0$>$} & \tiny{I'll be late}\\[-7 pt]
        \tiny{$<$Person1$>$} & \tiny{ok}\\[-7 pt]
        \tiny{$<$Person1$>$} & \tiny{do you want me to order something for you?}\\[-7 pt]
        \tiny{$<$Person1$>$} & \tiny{pasta?}\\[-7 pt]
        \tiny{$<$Person0$>$} & \tiny{Yes}\\[-7 pt]
        \tiny{$<$Person1$>$} & \tiny{with salmon?}\\[-7 pt]
        \tiny{$<$Person0$>$} & \tiny{Yes}\\[-7 pt]
        \tiny{$<$Person1$>$} & \tiny{Ok}\\[-7 pt]
        \tiny{$<$Person1$>$} & \tiny{how about basil?}\\[-7 pt]
        \tiny{$<$Person1$>$} & \tiny{Yes please!}\\
   \end{tabular}
\\

   \hline
\end{tabular}
\caption{Multiple conversations generated by the CN-Gen approach grounded on the same summary}
\label{table:cn_gen_samples}
\end{table}

A summary and a conversation from the Samsum dataset \cite{gliwa2019samsum}, along with the conversations generated by the three aforementioned algorithms are shown in Figure \ref{fig:example_convs}. More examples are provided in the Section B of the Appendix.

\section{Experiments}
We experiment on the Samsum \cite{gliwa2019samsum} dataset, which, to the best of our knowledge, is the only public large-scale conversation summarization dataset. We pre-process the dataset by replacing the personal names (ex: {\small John}) with unique tags (ex:{\small  $<$person\_$0>$}). First, we evaluate of the quality of generated conversations using automatic measures and human judgments, and then assess the performance of the generated conversations in a downstream summarization task after augmentation.
\subsection{Quality of the generated conversations}\label{sec:conversation_eval}
We evaluate the quality of the conversations generated by the three approaches that were introduced in Section \ref{sec:methods}. In Table \ref{tab:prop_dialogs} we show the properties of generated conversations and the ground truth conversations in the test set of Samsum dataset.

\begin{table}[t]
\centering
\resizebox{0.45\textwidth}{!}{
\begin{tabular}{l|ll}
\hline
Model        & Ave. Turns & Ave. Tokens/Turn \\\hline
Ground truth & $11.55\pm6.48$   & $7.10\pm6.29$           \\\hline
SL-Conv-Gen  & $10.54\pm6.80$   & $5.69\pm4.40$           \\
RL-Conv-Gen  & $8.40\pm4.78$    & $5.14\pm3.64$           \\
CN-Conv-Gen  & $9.70\pm5.67$    & $5.62\pm4.05$          \\\hline
\end{tabular}}
\caption{Properties of the generated conversations.}
\label{tab:prop_dialogs}
\end{table}

\textbf{Automatic Evaluation: }
We trained the conversation generation models on the Samsum training set and generated conversations on the test set.  We compare the generated conversation with the ground truth conversations using the measures used by \citet{sharma2017nlgeval} to evaluate conversation system responses. The results shown in Table \ref{tab:auto_eval_dialogs} suggest that CN-Gen outperform the SL-Gen and RL-Gen on all measures.



\begin{table*}[ht]
\begin{minipage}[t]{0.45\linewidth}
\centering
    \resizebox{0.97\textwidth}{!}{%
		\begin{tabular}{l|lll}
        \hline
        Model       & BLEU-4 & METEOR & ROUGE-L \\\hline
        SL-Gen & 2.81  & 12.06  & 21.53    \\
        RL-Gen & 3.53  & 12.29  & 25.40   \\
        CN-Gen & \textbf{4.94}  & \textbf{15.64}  & \textbf{26.22}   \\\hline
        \end{tabular}}
        \caption{Evaluation of generated conversations against ground truth conversations}
        \label{tab:auto_eval_dialogs}
\end{minipage}\hfill
\begin{minipage}[t]{0.45\linewidth}
\centering
\resizebox{0.99\textwidth}{!}{
\begin{tabular}{l|lll}
\hline
Model       & ROUGE\_1 & ROUGE\_2 & ROUGE\_L \\
\hline
SL-Gen & 46.85    & 25.29    & 45.97    \\
RL-Gen & 52.51    & 31.23    & 51.68    \\
CN-Gen & \textbf{53.46}    & \textbf{32.52}    &\textbf{52.93} \\
\hline
\end{tabular}}
\caption{Rouge F1 evaluation of summaries of conversations against the ground truth summaries}
\label{tab:auto_eval_summary}
\end{minipage}\hfill
\end{table*}

\begin{table*}[ht]
\begin{minipage}[t]{0.45\linewidth}
\centering
    \resizebox{0.75\textwidth}{!}{%
    
    \begin{tabular}{l|ccl}
\hline
Model & Info & Gram & Cohe \\ \hline
{\small Ground-Truth} & 4.56 & 4.46 & 4.47 \\ \hline
SL-Gen & 2.22 & 2.85 & 2.37 \\
RL-Gen & \textbf{3.20} & \textbf{3.50} & \textbf{3.14} \\
CN-Gen & 3.10 & 3.43 & 3.09 \\ \hline
\end{tabular}}
\caption{Human evaluation of generated conversations}
\label{tab:human-eval-dialog}
  \end{minipage}\hfill
\begin{minipage}[t]{0.45\linewidth}
\centering
\resizebox{0.75\textwidth}{!}{
\begin{tabular}{l|ccc}
\hline
Model       & Info & Gram & Cohe \\
\hline
{\small Ground-Truth}  & 0.04 & 0.22   & 0.25        \\
\hline
SL-Gen  & 0.35  & 0.26 & 0.42         \\
RL-Gen & 0.47   & 0.35  & 0.45        \\
CN-Gen  &0.60  & 0.40  & 0.60     \\
\hline
\end{tabular}}
\caption{Average Cohen's Kappa for human evaluation of generated conversations}
\label{tab:kappa}
\end{minipage}\hfill
\end{table*}

    

We also compare the summaries of generated conversations (generated by the Summary Generator) with the ground truth summaries, and the results are shown in Table \ref{tab:auto_eval_summary}. We believe that this is a semantic evaluation of the conversations, as the summaries capture the crux of the conversations. According to the results, CN-Gen outperforms the other two methods. This, along with the previous result suggest that the conversations produced by CN-Gen are the most similar to the ground truth conversations. 

\textbf{Human Evaluation: } 
To evaluate the quality of generated conversations, we randomly selected $50$ summaries  from the Samsum test dataset and generated conversations using the three models. 
Three NLP experts were then asked to read the ground truth summary and rank the four conversations (3 generated and the ground truth conversation) using a [1-5] scale according to Grammaticality, Coherency, and Informativeness, with respect to the ground truth summary.
Results are shown in table~\ref{tab:human-eval-dialog}. As expected, the ground-truth conversations obtained the highest scores on all three aspects and can be considered as an upper bound for this task. RL-Gen and CN-Gen obtained higher scores than SL-Gen and relatively good scores compared to the Ground Truth conversations. This corroborates the assumption that our proposed models generate high quality conversations. The Welch Two Sample t-test \cite{welch1947generalization} shows that both RL-Gen and CN-Gen models outperform the SL-Gen model statistically significantly with $p<0.0001$. However, there is no statistical significance between the results obtained from RL-Gen and CN-Gen. We report  in Table~\ref{tab:kappa} the average quadratic Cohen's Kappa calculated over the three possible combinations of two judges \cite{toledo2019automatic}.


CN-Gen obtained the best scores during the automatic evaluation, while RL-Gen got the best scores from the human evaluation. The CN-Gen conversations are longer than the RL-Gen conversation by $1.3$ turns on average (see Table \ref{tab:prop_dialogs}), and hence would contain more word overlap with the ground truth. This results in better automatic evaluation scores for the CN-Gen, while the humans prefer short targeted  conversations generated by RL-Gen.


\subsection{Evaluation on the summarization task}
To further evaluate the quality of the generate conversations, we augmented a conversation summarization dataset with generated conversations and evaluated the summarization model. We followed the following process: (1) We randomly selected x\% of the summaries of the dataset and trained our conversation generation models, (2) The trained models were applied on the other (y=100-x\%) of the summaries and generated conversations, (3) Those generated conversations along with the original summaries were added to the data. Using this approach, we can add extra y\% (summary, conversation) pairs to the training data, (4) The conversation summarization model (discussed in Section \ref{sec:methods} under `Summary Generator`) was trained on the augmented data. We compare the performance of the conversation summarization model on the original dataset and with augmentation. 

\textbf{Automatic Evaluation: } 
We compare the three conversation generation methods at different augmentation percentages, and the results are shown in Table \ref{tab:results-overall}. 
\begin{table}
\centering
\resizebox{0.5\textwidth}{!}{
\begin{tabular}{l|l|ccc}
\hline
\multicolumn{1}{l|}{\multirow{2}{*}{Method}} & Augmentation \% & ROUGE\_1 & ROUGE\_2 & ROUGE\_L \\ \cline{2-5} 
\multicolumn{1}{l|}{} & 0\% (Original) & 51.84 & 30.98 & 43.98 \\ \hline
\multirow{5}{*}{SL-Gen} 
 & 10\% & 52.82 & 31.99 & 44.89 \\ 
 & 20\% & \textbf{52.90} & \textbf{32.01} & 44.97 \\
 & 30\% & 52.88 & 32.02 & \textbf{45.01} \\
 & 40\% & 52.61 & 31.98 & 44.96 \\
 & 50\% & 52.55 & 31.98 & 44.80 \\\hline
\multirow{5}{*}{RL-Gen} 
 & 10\% & 52.93 & 32.05 & 44.92 \\ 
 & 20\% & 53.30 & 32.15 & 45.20 \\
 & 30\% & \textbf{53.81} & \textbf{32.21} & \textbf{45.77} \\
 & 40\% & 52.86 & 32.06 & 44.99 \\
 & 50\% & 52.64 & 32.07 & 44.88 \\\hline
\multirow{5}{*}{CN-Gen} 
 & 10\% & 53.29 & 32.36 & 45.08 \\ 
 & 20\% & 53.36 & 32.53 & 45.27 \\
  & 30\% & \textbf{54.02} & \textbf{33.28} & \textbf{46.06} \\
  & 40\% & 52.14 & 31.76 & 44.14 \\
  & 50\% & 52.36 & 31.75 & 44.85 \\\hline
\end{tabular}
}
\caption{ROUGE F-1 evaluation on Samsum test set.}
\vspace{-0.4cm}
\label{tab:results-overall}
\end{table}
At all augmentation levels, the summarization models trained with augmented data outperform the summarization model trained on the original dataset (without augmentation). CN-Gen based augmentation produces the best accuracy compared to other two methods. One prevalent pattern is that, when augmentation data increases, the accuracy seems to increase up to a certain point and then starts to decrease. The best accuracies were found around 30\% data augmentation. We believe that more augmentation leads performance to drop due to the following reason. For augmenting with more data, we are left with less data to train the model for conversation generation (for 10\% augmentation, the conversation generation models are trained on 90\% of the data, while for 50\% augmentation, the models are trained only on 50\% of the data). Therefore as the augmentation increases, the quality of generated conversations go down. This leads to overall smaller gains in the summarization task with increased augmentation after some point. To neutralize the effect of increasing the data points during augmentation, we experimented with a baseline which over-samples the original training data at different percentages to obtain same number of training instances as the augmented datasets. While the ROUGE-2 obtained with the original training data is 30.98, oversampling at 10\%, 20\%, 30\%, 40\% and 50\%, only changes the ROUGE-2 to 30.55, 30.38, 30.74, 30.99 and 30.27 respectively. Hence, this suggests that oversampling hardly changes ROUGE scores obtained by training with the original dataset, while the augmentation according to our algorithms show significantly improved scores (as shown in Table \ref{tab:results-overall}).

\textbf{Human Evaluation: }
We recruited 3 NLP experts to evaluate 50 instances of summaries generated with data augmentation (RL-Gen, CN-Gen), and respective summaries generated without augmentation (No-Aug). Here we consider two aspects with respect to a ground-truth summary: Coherency (whether the summary is easy to read) and Focus (whether the summary represents the ground-truth summary). Following \cite{amplayo2020unsupervised} we use the Best-Worst Scaling method.
The score of each system is computed as the percentage of times it was chosen as the Best system minus times it was chosen as Worst. 
On the Coherency question,  RL-Gen, CN-Gen and No-Aug obtained scores of 12.6, 6.6 and -4.0 respectively. On the Focus question RL-Gen, CN-Gen, and No-Aug obtained scores of 14.6, 6.0 and -2.6 respectively. These results confirm that the use of augmentation improves the quality of the summaries. 

\section{Conclusion}

We investigated how the PLMs can be utilized to generate entire conversations that are grounded on a summary. We propose three approaches for conversation generation: SL-Gen, RL-Gen and CN-Gen and conducted multiple automatic and human evaluations to assess the quality of the generated conversations. Both automatic and human evaluations show that when compared to the ground truth conversations, RL-Gen and CN-Gen obtain high scores, suggesting that the proposed models generate high quality conversations. When a conversation  summarization  dataset is  augmented with the generated conversations, the performance of conversation summarization is improved (over to 7\% improvement in ROUGE-2 F-1), which also suggests that the proposed methods generate high quality conversations. 

\section{Ethics}
We have used the publicly available Samsum dataset (\url{https://huggingface.co/datasets/samsum}). For the human evaluation of both conversations and summaries, we recruited 3 NLP researchers, who have graduate degree in NLP and Machine Learning. The annotation task itself was executed on Appen.com platform. Before the official annotation, we sampled 10 tasks to get an estimate of the duration of the task, and to make sure the instructions are clear enough.

\bibliographystyle{acl_natbib}
\bibliography{anthology,acl2021}

\begin{thebibliography}{28}
\expandafter\ifx\csname natexlab\endcsname\relax\def\natexlab#1{#1}\fi

\bibitem[{Amplayo and Lapata(2020)}]{amplayo2020unsupervised}
Reinald~Kim Amplayo and Mirella Lapata. 2020.
\newblock Unsupervised opinion summarization with noising and denoising.
\newblock In \emph{Proceedings of the 58th Annual Meeting of the Association
  for Computational Linguistics}, pages 1934--1945.

\bibitem[{Anaby-Tavor et~al.(2020)Anaby-Tavor, Carmeli, Goldbraich, Kantor,
  Kour, Shlomov, Tepper, and Zwerdling}]{anaby2020not}
Ateret Anaby-Tavor, Boaz Carmeli, Esther Goldbraich, Amir Kantor, George Kour,
  Segev Shlomov, Naama Tepper, and Naama Zwerdling. 2020.
\newblock Do not have enough data? deep learning to the rescue!
\newblock In \emph{Proceedings of the AAAI Conference on Artificial
  Intelligence}, volume~34, pages 7383--7390.

\bibitem[{Brown et~al.(2020)Brown, Mann, Ryder, Subbiah, Kaplan, Dhariwal,
  Neelakantan, Shyam, Sastry, Askell et~al.}]{brown2020language}
Tom~B Brown, Benjamin Mann, Nick Ryder, Melanie Subbiah, Jared Kaplan, Prafulla
  Dhariwal, Arvind Neelakantan, Pranav Shyam, Girish Sastry, Amanda Askell,
  et~al. 2020.
\newblock Language models are few-shot learners.
\newblock \emph{arXiv preprint arXiv:2005.14165}.

\bibitem[{Budzianowski and Vuli{\'c}(2019)}]{budzianowski2019hello}
Pawe{\l} Budzianowski and Ivan Vuli{\'c}. 2019.
\newblock Hello, it’s gpt-2-how can i help you? towards the use of pretrained
  language models for task-oriented dialogue systems.
\newblock In \emph{Proceedings of the 3rd Workshop on Neural Generation and
  Translation}, pages 15--22.

\bibitem[{Budzianowski et~al.(2018)Budzianowski, Wen, Tseng, Casanueva, Ultes,
  Ramadan, and Gasic}]{budzianowski2018multiwoz}
Pawe{\l} Budzianowski, Tsung-Hsien Wen, Bo-Hsiang Tseng, I{\~n}igo Casanueva,
  Stefan Ultes, Osman Ramadan, and Milica Gasic. 2018.
\newblock Multiwoz-a large-scale multi-domain wizard-of-oz dataset for
  task-oriented dialogue modelling.
\newblock In \emph{Proceedings of the 2018 Conference on Empirical Methods in
  Natural Language Processing}, pages 5016--5026.

\bibitem[{Cachola et~al.(2020)Cachola, Lo, Cohan, and Weld}]{cachola2020tldr}
Isabel Cachola, Kyle Lo, Arman Cohan, and Daniel~S Weld. 2020.
\newblock Tldr: Extreme summarization of scientific documents.
\newblock In \emph{Proceedings of the 2020 Conference on Empirical Methods in
  Natural Language Processing: Findings}, pages 4766--4777.

\bibitem[{Gliwa et~al.(2019)Gliwa, Mochol, Biesek, and Wawer}]{gliwa2019samsum}
Bogdan Gliwa, Iwona Mochol, Maciej Biesek, and Aleksander Wawer. 2019.
\newblock Samsum corpus: A human-annotated dialogue dataset for abstractive
  summarization.
\newblock \emph{EMNLP-IJCNLP 2019}, page~70.

\bibitem[{Hardalov et~al.(2018)Hardalov, Koychev, and
  Nakov}]{hardalov2018towards}
Momchil Hardalov, Ivan Koychev, and Preslav Nakov. 2018.
\newblock Towards automated customer support.
\newblock In \emph{International Conference on Artificial Intelligence:
  Methodology, Systems, and Applications}, pages 48--59. Springer.

\bibitem[{Kang et~al.(2018)Kang, Zhang, Kummerfeld, Tang, and
  Mars}]{kang2018data}
Yiping Kang, Yunqi Zhang, Jonathan~K Kummerfeld, Lingjia Tang, and Jason Mars.
  2018.
\newblock Data collection for dialogue system: A startup perspective.
\newblock In \emph{Proceedings of the 2018 Conference of the North American
  Chapter of the Association for Computational Linguistics: Human Language
  Technologies, Volume 3 (Industry Papers)}, pages 33--40.

\bibitem[{Lewis et~al.(2019)Lewis, Liu, Goyal, Ghazvininejad, Mohamed, Levy,
  Stoyanov, and Zettlemoyer}]{lewis2019bart}
Mike Lewis, Yinhan Liu, Naman Goyal, Marjan Ghazvininejad, Abdelrahman Mohamed,
  Omer Levy, Ves Stoyanov, and Luke Zettlemoyer. 2019.
\newblock Bart: Denoising sequence-to-sequence pre-training for natural
  language generation, translation, and comprehension.
\newblock \emph{arXiv preprint arXiv:1910.13461}.

\bibitem[{Lewis et~al.(2020)Lewis, Liu, Goyal, Ghazvininejad, Mohamed, Levy,
  Stoyanov, and Zettlemoyer}]{lewis2020bart}
Mike Lewis, Yinhan Liu, Naman Goyal, Marjan Ghazvininejad, Abdelrahman Mohamed,
  Omer Levy, Veselin Stoyanov, and Luke Zettlemoyer. 2020.
\newblock Bart: Denoising sequence-to-sequence pre-training for natural
  language generation, translation, and comprehension.
\newblock In \emph{Proceedings of the 58th Annual Meeting of the Association
  for Computational Linguistics}, pages 7871--7880.

\bibitem[{Li et~al.(2016)Li, Monroe, Ritter, Jurafsky, Galley, and
  Gao}]{li2016deep}
Jiwei Li, Will Monroe, Alan Ritter, Dan Jurafsky, Michel Galley, and Jianfeng
  Gao. 2016.
\newblock Deep reinforcement learning for dialogue generation.
\newblock In \emph{Proceedings of the 2016 Conference on Empirical Methods in
  Natural Language Processing}, pages 1192--1202.

\bibitem[{Lowe et~al.(2015)Lowe, Pow, Serban, and Pineau}]{lowe2015ubuntu}
Ryan Lowe, Nissan Pow, Iulian~Vlad Serban, and Joelle Pineau. 2015.
\newblock The ubuntu dialogue corpus: A large dataset for research in
  unstructured multi-turn dialogue systems.
\newblock In \emph{Proceedings of the 16th Annual Meeting of the Special
  Interest Group on Discourse and Dialogue}, pages 285--294.

\bibitem[{McCowan et~al.(2005)McCowan, Carletta, Kraaij, Ashby, Bourban, Flynn,
  Guillemot, Hain, Kadlec, Karaiskos et~al.}]{mccowan2005ami}
Iain McCowan, Jean Carletta, Wessel Kraaij, Simone Ashby, S~Bourban, M~Flynn,
  M~Guillemot, Thomas Hain, J~Kadlec, Vasilis Karaiskos, et~al. 2005.
\newblock The ami meeting corpus.
\newblock In \emph{Proceedings of the 5th International Conference on Methods
  and Techniques in Behavioral Research}, volume~88, page 100. Citeseer.

\bibitem[{Min et~al.(2020)Min, Michael, Hajishirzi, and
  Zettlemoyer}]{min2020ambigqa}
Sewon Min, Julian Michael, Hannaneh Hajishirzi, and Luke Zettlemoyer. 2020.
\newblock Ambigqa: Answering ambiguous open-domain questions.
\newblock In \emph{Proceedings of the 2020 Conference on Empirical Methods in
  Natural Language Processing (EMNLP)}, pages 5783--5797.

\bibitem[{Mohapatra et~al.(2020)Mohapatra, Pandey, Contractor, and
  Joshi}]{mohapatra2020simulated}
Biswesh Mohapatra, Gaurav Pandey, Danish Contractor, and Sachindra Joshi. 2020.
\newblock Simulated chats for task-oriented dialog: Learning to generate
  conversations from instructions.
\newblock \emph{arXiv preprint arXiv:2010.10216}.

\bibitem[{Narayan et~al.(2018)Narayan, Cohen, and Lapata}]{narayan2018don}
Shashi Narayan, Shay~B Cohen, and Mirella Lapata. 2018.
\newblock Don’t give me the details, just the summary! topic-aware
  convolutional neural networks for extreme summarization.
\newblock In \emph{Proceedings of the 2018 Conference on Empirical Methods in
  Natural Language Processing}, pages 1797--1807.

\bibitem[{Paulus et~al.(2018)Paulus, Xiong, and Socher}]{paulus2018deep}
Romain Paulus, Caiming Xiong, and Richard Socher. 2018.
\newblock A deep reinforced model for abstractive summarization.
\newblock In \emph{International Conference on Learning Representations}.

\bibitem[{Raffel et~al.(2020)Raffel, Shazeer, Roberts, Lee, Narang, Matena,
  Zhou, Li, and Liu}]{raffel2020exploring}
Colin Raffel, Noam Shazeer, Adam Roberts, Katherine Lee, Sharan Narang, Michael
  Matena, Yanqi Zhou, Wei Li, and Peter~J Liu. 2020.
\newblock Exploring the limits of transfer learning with a unified text-to-text
  transformer.
\newblock \emph{Journal of Machine Learning Research}, 21(140):1--67.

\bibitem[{Rameshkumar and Bailey(2020)}]{rameshkumar2020storytelling}
Revanth Rameshkumar and Peter Bailey. 2020.
\newblock Storytelling with dialogue: A critical role dungeons and dragons
  dataset.
\newblock In \emph{Proceedings of the 58th Annual Meeting of the Association
  for Computational Linguistics}, pages 5121--5134.

\bibitem[{Schulman et~al.(2017)Schulman, Wolski, Dhariwal, Radford, and
  Klimov}]{schulman2017proximal}
John Schulman, Filip Wolski, Prafulla Dhariwal, Alec Radford, and Oleg Klimov.
  2017.
\newblock Proximal policy optimization algorithms.
\newblock \emph{arXiv preprint arXiv:1707.06347}.

\bibitem[{Sharma et~al.(2017)Sharma, El~Asri, Schulz, and
  Zumer}]{sharma2017nlgeval}
Shikhar Sharma, Layla El~Asri, Hannes Schulz, and Jeremie Zumer. 2017.
\newblock \href {http://arxiv.org/abs/1706.09799} {Relevance of unsupervised
  metrics in task-oriented dialogue for evaluating natural language
  generation}.
\newblock \emph{CoRR}, abs/1706.09799.

\bibitem[{Toledo et~al.(2019)Toledo, Gretz, Cohen-Karlik, Friedman, Venezian,
  Lahav, Jacovi, Aharonov, and Slonim}]{toledo2019automatic}
Assaf Toledo, Shai Gretz, Edo Cohen-Karlik, Roni Friedman, Elad Venezian, Dan
  Lahav, Michal Jacovi, Ranit Aharonov, and Noam Slonim. 2019.
\newblock Automatic argument quality assessment-new datasets and methods.
\newblock In \emph{Proceedings of the 2019 Conference on Empirical Methods in
  Natural Language Processing and the 9th International Joint Conference on
  Natural Language Processing (EMNLP-IJCNLP)}, pages 5629--5639.

\bibitem[{Welch(1947)}]{welch1947generalization}
Bernard~L Welch. 1947.
\newblock The generalization ofstudent's' problem when several different
  population variances are involved.
\newblock \emph{Biometrika}, 34(1/2):28--35.

\bibitem[{Wolf et~al.(2019)Wolf, Debut, Sanh, Chaumond, Delangue, Moi, Cistac,
  Rault, Louf, Funtowicz et~al.}]{wolf2019huggingface}
Thomas Wolf, Lysandre Debut, Victor Sanh, Julien Chaumond, Clement Delangue,
  Anthony Moi, Pierric Cistac, Tim Rault, R{\'e}mi Louf, Morgan Funtowicz,
  et~al. 2019.
\newblock Huggingface's transformers: State-of-the-art natural language
  processing.
\newblock \emph{ArXiv}, pages arXiv--1910.

\bibitem[{Wu et~al.(2020)Wu, Li, and Yu}]{wu2020textgail}
Qingyang Wu, Lei Li, and Zhou Yu. 2020.
\newblock Textgail: Generative adversarial imitation learning for text
  generation.
\newblock \emph{arXiv preprint arXiv:2004.13796}.

\bibitem[{Yang et~al.(2020)Yang, Malaviya, Fernandez, Swayamdipta, Le~Bras,
  Wang, Bhagavatula, Choi, and Downey}]{yang2020g}
Yiben Yang, Chaitanya Malaviya, Jared Fernandez, Swabha Swayamdipta, Ronan
  Le~Bras, Ji-Ping Wang, Chandra Bhagavatula, Yejin Choi, and Doug Downey.
  2020.
\newblock G-daug: Generative data augmentation for commonsense reasoning.
\newblock In \emph{Proceedings of the 2020 Conference on Empirical Methods in
  Natural Language Processing: Findings}, pages 1008--1025.

\bibitem[{Zhang et~al.(2018)Zhang, Dinan, Urbanek, Szlam, Kiela, and
  Weston}]{zhang2018personalizing}
Saizheng Zhang, Emily Dinan, Jack Urbanek, Arthur Szlam, Douwe Kiela, and Jason
  Weston. 2018.
\newblock Personalizing dialogue agents: I have a dog, do you have pets too?
\newblock In \emph{Proceedings of the 56th Annual Meeting of the Association
  for Computational Linguistics (Volume 1: Long Papers)}, pages 2204--2213.

\end{thebibliography}

\appendix

\section{Model Training and Hyperparameter Details}

\subsection{Supervised Conversation Generation (SL-Conv-Gen)}

We fine-tune a GPT-2 language model using the implementation available at HuggingFace \cite{wolf2019huggingface}. The hyper-parameters used during training and inference are shown below. The model takes around 6 hours to train on 2 V100 GPUs (single machine).

\begin{Verbatim}[fontsize=\small]

model_name_or_path: gpt2
per_gpu_train_batch_size: 4
per_gpu_eval_batch_size: 4
gradient_accumulation_steps: 4
learning_rate: 6.25e-5
adam_epsilon: 1e-8
max_grad_norm: 1.0
num_train_epochs: 10
warmup_steps: 500
min_length: 20
max_length: 512
top_k: 0
top_p: 0.95

\end{Verbatim}

\subsection{Summary Generator}
 We use DistilBART instance\footnote{\url{https://huggingface.co/sshleifer/distilbart-cnn-12-6}} fine-tuned on the extreme summarization (XSum) task, and we fine-tune this model further on the Samsum dataset. The model takes around 12 hours to train on 2 V100 GPUs (single machine).

The hyperparameters used for training the DistilBART model are as follows:

\begin{Verbatim}[fontsize=\small]
train_batch_size: 4
eval_batch_size: 4
num_train_epochs: 10
model_name_or_path: sshleifer/distilbart
-xsum-12-6
learning_rate: 3e-5
val_check_interval: 0.1
max_source_length: 512
max_target_length: 80
\end{Verbatim}

\subsection{Reinforced Learning based conversation generation (RL-Conv-Gen)}

To train the RL based conversation generation model, we adapted a publicly available Proximal Policy Optimization (PPO) implementation \footnote{
\url{https://github.com/lvwerra/trl}
}. The model takes around 12 hours to train on 2 V100 GPUs (single machine).
Following hyper-parameters were used to train the model.
\begin{Verbatim}[fontsize=\small]
steps: 10000
batch_size: 16
forward_batch_size: 4
learning_rate: 1.41e-5
init_kl_coef:0.2
target: 6
horizon:10000
gamma:1
lam:0.95
cliprange: 0.2
cliprange_value: 0.2
vf_coef: 0.1
\end{Verbatim}

\section{Sample summaries with corresponding ground-truth}

Figure~\ref{figure:sample} shows some samples of dialogs with their corresponding summaries - ground-truth and automatic generated ones.

\begin{figure*}[t]
\centering
\begin{tabular*}{1\textwidth}{p{100pt}|p{100pt}|p{100pt}|p{100pt}}
   \hline
    \multicolumn{4}{l}{\tiny{\textbf{Summary:} Person0 closed some deals today. Person1 didn't manage to do it.}}\\
   \hline
   
   \begin{tabular}[t]{p{.21\linewidth}p{.64\linewidth}}
   \multicolumn{2}{l}{\tiny{\textbf{Ground Truth Dialog:}}}\\[-4 pt]
        \tiny{$<$Person0$>$} & \tiny{It's been a long day.}\\[-7 pt]
        \tiny{$<$Person1$>$} & \tiny{Same here, Person0, good or bad?}\\[-7 pt]
        \tiny{$<$Person0$>$} & \tiny{Very busy. I closed some deals but I had a lot of stress also.}\\[-7 pt]
        \tiny{$<$Person1$>$} & \tiny{Lucky you. I couldn't close any deal. It annoys me. Some customers are negotiating for days, weeks.}\\[-7 pt]
        \tiny{$<$Person0$>$} & \tiny{Maybe they're just fishing. Want to compare offers.}\\[-7 pt]
        \tiny{$<$Person1$>$} & \tiny{That's what they do.}
   \end{tabular}

   & 
   \begin{tabular}[t]{p{.21\linewidth}p{.64\linewidth}}
   \multicolumn{2}{l}{\tiny{\textbf{SL-Gen:}}}\\[-4 pt]
        \tiny{$<$Person0$>$} & \tiny{Hi, are you still alive?}\\[-7 pt]
        \tiny{$<$Person1$>$} & \tiny{Hi, yes, still alive.}\\[-7 pt]
        \tiny{$<$Person0$>$} & \tiny{I closed some deals today.}\\[-7 pt]
        \tiny{$<$Person1$>$} & \tiny{Yeah, that's awesome.}\\[-7 pt]
        \tiny{$<$Person0$>$} & \tiny{Thanks.}\\[-7 pt]
   \end{tabular}

    & 


   \begin{tabular}[t]{p{.21\linewidth}p{.64\linewidth}}
   \multicolumn{2}{l}{\tiny{\textbf{RL-Gen:}}}\\[-4 pt]
        \tiny{$<$Person0$>$} & \tiny{Hey, how are you?}\\[-7 pt]
        \tiny{$<$Person1$>$} & \tiny{Good, busy as usual.}\\[-7 pt]
        \tiny{$<$Person0$>$} & \tiny{I closed some deals today.}\\[-7 pt]
        \tiny{$<$Person1$>$} & \tiny{Too late, didn't manage to do it.}\\[-7 pt]
        \tiny{$<$Person0$>$} & \tiny{Sorry.}\\[-7 pt]
   \end{tabular}

   &
   
   \begin{tabular}[t]{p{.21\linewidth}p{.64\linewidth}}
   \multicolumn{2}{l}{\tiny{\textbf{CN-Gen:}}}\\[-4 pt]
        \tiny{$<$Person0$>$} & \tiny{Did you see this? <file\_other>}\\[-7 pt]
        \tiny{$<$Person1$>$} & \tiny{Cool. What's up?}\\[-7 pt]
        \tiny{$<$Person0$>$} & \tiny{I closed some deals today and I feel good about it.}\\[-7 pt]
        \tiny{$<$Person1$>$} & \tiny{Great. I didn't manage to do it today. I'm sorry. I had no choice.}\\[-7 pt]
        \tiny{$<$Person0$>$} & \tiny{I'm sorry.}\\[-7 pt]
        \tiny{$<$Person1$>$} & \tiny{I didn't have the tools.}\\[-7 pt]
   \end{tabular}
\\

   \hline
   \hline

    \multicolumn{4}{l}{\tiny{\textbf{Summary:} Person0 bought a table, six chairs, a vase and a pile of clothes and the second hand shop downtown. She paid 70 euros for everything.}}\\
   \hline
   
   \begin{tabular}[t]{p{.21\linewidth}p{.64\linewidth}}
   \multicolumn{2}{l}{\tiny{\textbf{Ground Truth Dialog:}}}\\[-4 pt]
        \tiny{$<$Person0$>$} & \tiny{I went to the second hand shop downtown.}\\[-7 pt]
        \tiny{$<$Person1$>$} & \tiny{Cool Person0  What did you get?}\\[-7 pt]
        \tiny{$<$Person0$>$} & \tiny{Lots of stuff. A table, six chairs, a vase, a pile of clothes}\\[-7 pt]
        \tiny{$<$Person1$>$} & \tiny{That's really a lot of stuff Person0. Send us pictures}\\[-7 pt]
        \tiny{$<$Person0$>$} & \tiny{<photo\_file>}\\[-7 pt]
        \tiny{$<$Person1$>$} & \tiny{Beautiful table.Is it wood?}\\[-7 pt]
        \tiny{$<$Person0$>$} & \tiny{Yes, oak.  }\\[-7 pt]
        \tiny{$<$Person1$>$} & \tiny{Awesome! These are the clothes? I love the black dress.}\\[-7 pt]
        \tiny{$<$Person0$>$} & \tiny{It's absolutely beautiful. As soon as I saw it I knew it would be mine. I'm so happy with what I bought.}\\[-7 pt]
   \end{tabular}

   & 
   \begin{tabular}[t]{p{.21\linewidth}p{.64\linewidth}}
   \multicolumn{2}{l}{\tiny{\textbf{SL-Gen:}}}\\[-4 pt]
        \tiny{$<$Person0$>$} & \tiny{I bought the second hand shop downtown.}\\[-7 pt]
        \tiny{$<$Person1$>$} & \tiny{So what happened?}\\[-7 pt]
        \tiny{$<$Person0$>$} & \tiny{I bought everything.}\\[-7 pt]
        \tiny{$<$Person1$>$} & \tiny{What do you mean?}\\[-7 pt]
        \tiny{$<$Person0$>$} & \tiny{I paid 70 euro for everything.}\\[-7 pt]
        \tiny{$<$Person1$>$} & \tiny{How much?}\\[-7 pt]
        \tiny{$<$Person0$>$} & \tiny{6 chairs, six chairs, a vase and a pile of clothes.}\\[-7 pt]
        \tiny{$<$Person1$>$} & \tiny{Sounds like a bargain.}\\[-7 pt]
        \tiny{$<$Person0$>$} & \tiny{ou should be in business.}\\[-7 pt]
   \end{tabular}

    & 


   \begin{tabular}[t]{p{.21\linewidth}p{.64\linewidth}}
   \multicolumn{2}{l}{\tiny{\textbf{RL-Gen:}}}\\[-4 pt]
        \tiny{$<$Person0$>$} & \tiny{Do you remember the second hand in downtown?}\\[-7 pt]
        \tiny{$<$Person1$>$} & \tiny{Yes, I do. What's inside?}\\[-7 pt]
        \tiny{$<$Person0$>$} & \tiny{A table, six chairs, a vase and a pile of clothes.}\\[-7 pt]
        \tiny{$<$Person1$>$} & \tiny{Is it expensive?}\\[-7 pt]
        \tiny{$<$Person0$>$} & \tiny{70 euros.}\\[-7 pt]
   \end{tabular}

   &
   
   \begin{tabular}[t]{p{.21\linewidth}p{.64\linewidth}}
   \multicolumn{2}{l}{\tiny{\textbf{CN-Gen:}}}\\[-4 pt]
        \tiny{$<$Person0$>$} & \tiny{Look what I've bought! A table, six chairs, a vase. I bought a table, six chairs, a vase and a pile of clothes.}\\[-7 pt]
        \tiny{$<$Person1$>$} & \tiny{How much did you pay for everything?}\\[-7 pt]
        \tiny{$<$Person0$>$} & \tiny{I paid 70 euro for everything. And I thought it would cost me a fortune! I don't even know how much.}\\[-7 pt]
        \tiny{$<$Person1$>$} & \tiny{I am glad.}\\[-7 pt]
        \tiny{$<$Person0$>$} & \tiny{Me too. 70 euro for everything, including a vase!!!}\\[-7 pt]
   \end{tabular}
\\

   \hline
   \hline
   
    \multicolumn{4}{l}{\tiny{\textbf{Summary:} Person1 is not at home. Person0 wants Person1 to keep her pasta in the microwave.}}\\
   \hline
   
   \begin{tabular}[t]{p{.21\linewidth}p{.64\linewidth}}
   \multicolumn{2}{l}{\tiny{\textbf{Ground Truth Dialog:}}}\\[-4 pt]
        \tiny{$<$Person0$>$} & \tiny{U both at home?}\\[-7 pt]
        \tiny{$<$Person1$>$} & \tiny{No. I've just got ur msg. Why did u ask about it?}\\[-7 pt]
        \tiny{$<$Person0$>$} & \tiny{No reason. Keep my pasta in the microwave.}\\[-7 pt]
        \tiny{$<$Person1$>$} & \tiny{I haven't cooked anything.}\\[-7 pt]
   \end{tabular}

   & 
   \begin{tabular}[t]{p{.21\linewidth}p{.64\linewidth}}
   \multicolumn{2}{l}{\tiny{\textbf{SL-Gen:}}}\\[-4 pt]
        \tiny{$<$Person0$>$} & \tiny{How are you doing?}\\[-7 pt]
        \tiny{$<$Person1$>$} & \tiny{I'm not at home}\\[-7 pt]
        \tiny{$<$Person0$>$} & \tiny{Why?}\\[-7 pt]
        \tiny{$<$Person1$>$} & \tiny{I wanted to keep some pasta in the microwave.}\\[-7 pt]
        \tiny{$<$Person0$>$} & \tiny{No need to.}\\[-7 pt]
        \tiny{$<$Person1$>$} & \tiny{In the microwave?}\\[-7 pt]
        \tiny{$<$Person0$>$} & \tiny{Yes.}\\[-7 pt]
   \end{tabular}

    & 


   \begin{tabular}[t]{p{.21\linewidth}p{.64\linewidth}}
   \multicolumn{2}{l}{\tiny{\textbf{RL-Gen:}}}\\[-4 pt]
        \tiny{$<$Person0$>$} & \tiny{Are you at home?}\\[-7 pt]
        \tiny{$<$Person1$>$} & \tiny{No, I'm still at work.}\\[-7 pt]
        \tiny{$<$Person0$>$} & \tiny{Good, I should be home in about 15 minutes.}\\[-7 pt]
        \tiny{$<$Person1$>$} & \tiny{Can you put my pasta in the microwave?}\\[-7 pt]
        \tiny{$<$Person0$>$} & \tiny{Sure, I'll check it.}\\[-7 pt]
        \tiny{$<$Person1$>$} & \tiny{Thanks!}\\[-7 pt]

   \end{tabular}

   &
   
   \begin{tabular}[t]{p{.21\linewidth}p{.64\linewidth}}
   \multicolumn{2}{l}{\tiny{\textbf{CN-Gen:}}}\\[-4 pt]
        \tiny{$<$Person0$>$} & \tiny{Are you at home?}\\[-7 pt]
        \tiny{$<$Person1$>$} & \tiny{No, I'm busy at work. What do you want?}\\[-7 pt]
        \tiny{$<$Person0$>$} & \tiny{I want to put some pasta in microwave.}\\[-7 pt]
        \tiny{$<$Person1$>$} & \tiny{OK. I'll try.}\\[-7 pt]
   \end{tabular}
  \\
   \hline
   \hline

    \multicolumn{4}{l}{\tiny{\textbf{Summary:} Person0 needs Person1's help as he cannot get the application running.}}\\
   \hline
   
  \begin{tabular}[t]{p{.21\linewidth}p{.64\linewidth}}
   \multicolumn{2}{l}{\tiny{\textbf{Ground Truth Dialog:}}}\\[-4 pt]
        \tiny{$<$Person0$>$} & \tiny{Person1, I need your help?}\\[-7 pt]
        \tiny{$<$Person1$>$} & \tiny{What's up, mate?}\\[-7 pt]
        \tiny{$<$Person0$>$} & \tiny{I can't get the application running.}\\[-7 pt]
        \tiny{$<$Person1$>$} & \tiny{Have you switched the computer on?}\\[-7 pt]
        \tiny{$<$Person0$>$} & \tiny{Very funny!!!!!!!!!!!!!}\\[-7 pt]
        \tiny{$<$Person1$>$} & \tiny{OK. Sorry. I can see it's serious.}\\[-7 pt]
        \tiny{$<$Person0$>$} & \tiny{Yeah, man. It is f**cking serious.}\\[-7 pt]
        \tiny{$<$Person1$>$} & \tiny{I'll be with you right now.}\\[-7 pt]
        \tiny{$<$Person0$>$} & \tiny{Thanks.}\\[-7 pt]
   \end{tabular}

   & 
   \begin{tabular}[t]{p{.21\linewidth}p{.64\linewidth}}
   \multicolumn{2}{l}{\tiny{\textbf{SL-Gen:}}}\\[-4 pt]
        \tiny{$<$Person0$>$} & \tiny{Hey! I have a problem with the application.}\\[-7 pt]
        \tiny{$<$Person1$>$} & \tiny{Hello, Person0}\\[-7 pt]
        \tiny{$<$Person0$>$} & \tiny{Could you please help me to get the application running? I cannot.}\\[-7 pt]
        \tiny{$<$Person1$>$} & \tiny{Why?}\\[-7 pt]
        \tiny{$<$Person0$>$} & \tiny{I have a problem with my laptop. The application is running on the USB port. I cannot access the computer. The computer is running. If I try to access the computer, the application will start. I hope that you can help me.}\\[-7 pt]
        \tiny{$<$Person1$>$} & \tiny{Okay.}\\[-7 pt]
        \tiny{$<$Person0$>$} & \tiny{Thank you, Person1.}\\[-7 pt]
   \end{tabular}

    & 


   \begin{tabular}[t]{p{.21\linewidth}p{.64\linewidth}}
   \multicolumn{2}{l}{\tiny{\textbf{RL-Gen:}}}\\[-4 pt]
        \tiny{$<$Person0$>$} & \tiny{Person1, you there?}\\[-7 pt]
        \tiny{$<$Person1$>$} & \tiny{Yeah, what's up?}\\[-7 pt]
        \tiny{$<$Person0$>$} & \tiny{Do you need my help?}\\[-7 pt]
        \tiny{$<$Person1$>$} & \tiny{What's wrong?}\\[-7 pt]
        \tiny{$<$Person0$>$} & \tiny{I can't get the application running.}\\[-7 pt]
        \tiny{$<$Person1$>$} & \tiny{Did you try charging it first?}\\[-7 pt]
        \tiny{$<$Person0$>$} & \tiny{Yes, I tried that, but it doesn't work.}\\[-7 pt]
        \tiny{$<$Person1$>$} & \tiny{Try refreshing the database.}\\[-7 pt]
        \tiny{$<$Person0$>$} & \tiny{Damn, thanks!}\\[-7 pt]
   \end{tabular}

   &
   
   \begin{tabular}[t]{p{.21\linewidth}p{.64\linewidth}}
   \multicolumn{2}{l}{\tiny{\textbf{CN-Gen:}}}\\[-4 pt]
        \tiny{$<$Person0$>$} & \tiny{I cannot get the application running.}\\[-7 pt]
        \tiny{$<$Person1$>$} & \tiny{What's up?}\\[-7 pt]
        \tiny{$<$Person0$>$} & \tiny{Can you get the application running?}\\[-7 pt]
        \tiny{$<$Person1$>$} & \tiny{You sure you can't? Try and click on it.}\\[-7 pt]
        \tiny{$<$Person0$>$} & \tiny{You're so funny, Person1.}\\[-7 pt]
        \tiny{$<$Person1$>$} & \tiny{OK. I'm coming.}\\[-7 pt]
        \tiny{$<$Person0$>$} & \tiny{Thanks.}\\[-7 pt]
   \end{tabular}
  \\
   \hline
   \hline

    \multicolumn{4}{l}{\tiny{\textbf{Summary:} Person0 and Person1 will meet the new person in an hour.}}\\
  \hline
   
  \begin{tabular}[t]{p{.21\linewidth}p{.64\linewidth}}
  \multicolumn{2}{l}{\tiny{\textbf{Ground Truth Dialog:}}}\\[-4 pt]
        \tiny{$<$Person0$>$} & \tiny{Can we meet with the new person soon?}\\[-7 pt]
        \tiny{$<$Person1$>$} & \tiny{Sure. In an hour okay?}\\[-7 pt]
        \tiny{$<$Person0$>$} & \tiny{Perfect.}\\[-7 pt]
  \end{tabular}

  & 
  \begin{tabular}[t]{p{.21\linewidth}p{.64\linewidth}}
  \multicolumn{2}{l}{\tiny{\textbf{SL-Gen:}}}\\[-4 pt]
        \tiny{$<$Person0$>$} & \tiny{Hi, babe.}\\[-7 pt]
        \tiny{$<$Person1$>$} & \tiny{Hi.}\\[-7 pt]
        \tiny{$<$Person0$>$} & \tiny{We're meeting in an hour?}\\[-7 pt]
        \tiny{$<$Person1$>$} & \tiny{I'd prefer to be there in an hour.}\\[-7 pt]
        \tiny{$<$Person0$>$} & \tiny{Great!}\\[-7 pt]
  \end{tabular}

    & 


  \begin{tabular}[t]{p{.21\linewidth}p{.64\linewidth}}
  \multicolumn{2}{l}{\tiny{\textbf{RL-Gen:}}}\\[-4 pt]
        \tiny{$<$Person0$>$} & \tiny{Did you get that message?}\\[-7 pt]
        \tiny{$<$Person1$>$} & \tiny{No, why?}\\[-7 pt]
        \tiny{$<$Person0$>$} & \tiny{I might have lost it.}\\[-7 pt]
        \tiny{$<$Person1$>$} & \tiny{Oh no, it's my mistake. I'm sorry.}\\[-7 pt]
        \tiny{$<$Person0$>$} & \tiny{Ok, don't worry. We can meet in an hour?}\\[-7 pt]
        \tiny{$<$Person1$>$} & \tiny{Ok!}\\[-7 pt]
  \end{tabular}

  &
   
  \begin{tabular}[t]{p{.21\linewidth}p{.64\linewidth}}
  \multicolumn{2}{l}{\tiny{\textbf{CN-Gen:}}}\\[-4 pt]
        \tiny{$<$Person0$>$} & \tiny{Wanna meet the new person?}\\[-7 pt]
        \tiny{$<$Person1$>$} & \tiny{Sure, I'll be there in an hour.}\\[-7 pt]
        \tiny{$<$Person0$>$} & \tiny{Perfect!}\\[-7 pt]
  \end{tabular}
  \\
  \hline

\end{tabular*}
\caption{Samples of dialogs with their corresponding summaries - ground-truth and automatic generated ones}
\label{figure:sample}
\end{figure*}

\end{document}